%% file: main.tex
\documentclass[conference]{IEEEtran}
\IEEEoverridecommandlockouts
\usepackage{cite}
\usepackage{amsmath,amssymb,amsfonts}
\usepackage{algorithmic}

\usepackage{algorithm} 
\usepackage{hyperref}
\usepackage{graphicx}
\usepackage{textcomp}
\usepackage{xcolor}
\def\BibTeX{{\rm B\kern-.05em{\sc i\kern-.025em b}\kern-.08em
    T\kern-.1667em\lower.7ex\hbox{E}\kern-.125emX}}
\begin{document}

\title{Bootstrap Advantage Estimation for \\ Policy Optimization in Reinforcement Learning}

\author{
\IEEEauthorblockN{Md Masudur Rahman, Yexiang Xue}
\IEEEauthorblockA{\textit{Department of Computer Science} \\
\textit{Purdue University, 
West Lafayette, Indiana, USA} \\
\{rahman64, yexiang\}@purdue.edu}
}

\maketitle

\input{paper}

\bibliographystyle{IEEEtran}
\bibliography{main}
\end{document}

%% file: paper.tex
\begin{abstract}
This paper proposes an advantage estimation approach based on data augmentation for policy optimization. Unlike using data augmentation on the input to learn value and policy function as existing methods use, our method uses data augmentation to compute a bootstrap advantage estimation. This Bootstrap Advantage Estimation (BAE) is then used for learning and updating the gradient of policy and value function. To demonstrate the effectiveness of our approach, we conducted experiments on several environments. These environments are from three benchmarks: Procgen, Deepmind Control, and Pybullet, which include both image and vector-based observations; discrete and continuous action spaces. We observe that our method reduces the policy and the value loss better than the Generalized advantage estimation (GAE) method and eventually improves cumulative return. Furthermore, our method performs better than two recently proposed data augmentation techniques (RAD and DRAC). Overall, our method performs better empirically than baselines in sample efficiency and generalization, where the agent is tested in unseen environments.
\end{abstract}


\input{intro.tex}
\input{background.tex}

\input{method.tex}
\input{experiment.tex}
\input{relatedwork.tex}
\input{conclusion.tex}

%% file: intro.tex
\section{Introduction}
The policy gradient method directly involves learning policy function, which enjoys performance improvement in function approximation settings \cite{sutton1999policy}. 
The policy gradient theorem gives a rather simple formulation of the gradient estimation, which gives an unbias estimation \cite{williams1992simple}. However, it requires the return estimation of the entire trajectory, leading to very high variance. A commonly used technique to reduce variance is to use a baseline, which can help reduce variance without introducing bias. Several effective methods originated from this concept \cite{schulman2017proximal,schulman2015trust}. An effective way is to use the value function as a baseline to indicate whether the action taken by the current policy is better than the average action taken in that state, which can be formulated as an advantage estimation.
However, based on a single trajectory, the estimate can be local and have high variance. Thus we can add a truncated scenario where the value function can potentially give a global estimate. Combining these two is the Generalized advantage estimation \cite{Schulmanetal_ICLR2016} which shows strong empirical results \cite{andrychowicz2021what}.

However, due to procedural style content generation, the value estimation can be erroneous and give different advantage estimates even when the observation context changes. At the same time, the semantic meaning remains the same. As the procedural scenario can exist in a real-world scenario and might cause agent to perform sup-optimally \cite{Song2020Observational, zhang2018study}, thus this advantage estimation can be problematic in those scenarios, resulting in poor sample efficiency. This issue might exist partly due to difficulty in reducing policy estimation (policy loss) and value function estimation (value loss).

This paper proposes Bootstrap Advantage Estimation (BAE), which calculates the advantage estimation by computing advantage estimation on the original and its transformed observations. We assume the transformation to be a semantic invariant; the reward semantic remains the same, but the contextual information can be changed. 
For example, if the background of a game is not relevant, then changing the background color from red to green can be a semantic invariant transformation. The ultimate goal is to train an agent to be robust against any such background change, which performs well in the blue background in this example.
The transformed observation can be of any form; we experimented with data augmentation-based observation transformation (e.g., random crop, amplitude scaling). The intuition is that taking advantage of estimate over augmented data forces the advantage estimate to consider the error over many variations of the observations. We demonstrate our BAE on the policy gradient method (i.e., PPO \cite{schulman2017proximal}) and show a comparison over GAE-based estimation. We observed that our method BAE achieved better sample efficiency and zero-shot generalization in Procgen environments (starpilot and miner) with image-based observation. 

In recent times, data augmentation demonstrated an effective choice in improving sample efficiency in high-dimensional observation space and improving generalization \cite{cobbe2019quantifying,laskin2020reinforcement,raileanu2020automatic}.
Though this process sometimes generates empirical success \cite{laskin2020reinforcement}, such methods might lead to detrimental performance \cite{raileanu2020automatic} as we observed in our experiments. To mitigate this issue, the DRAC \cite{raileanu2020automatic} method suggests regularizing the policy network and value network by augmented observation and not using augmented data for policy training. In contrast, we propose a novel way to leverage data augmentation. Our method augmented observations for advantage estimation, one of the core components of many policy optimization algorithms (e.g., PPO). We conducted extensive experiments on six environments consisting of image and vector-based observation; and discrete and continuous action spaces. Our method falls in the general model-free on-policy category, and we experimented with Proximal Policy Optimization (PPO) \cite{schulman2017proximal} in this paper.
In particular, our experiments on Procgen Starpilot and Miner environments demonstrate that our method can be beneficial in the zero-shot generalization setup compared to baseline GAE \cite{Schulmanetal_ICLR2016}, and two data augmentation techniques: RAD\cite{laskin2020reinforcement}, and DRAC \cite{raileanu2020automatic}. 

We further evaluated our method on several robotic locomotion tasks with the high-dimensional observation from Deepmind Control Suite \cite{tunyasuvunakool2020}: Quadruped Run, and Cartpole - Three Poles; and PyBullet\cite{coumans2021}: Minitaur and HalfCheetah.

In experiments, we observe that our method BAE performs better than baseline agents, including base PPO, RAD, and DRAC. Our method achieves a much lower loss for policy and value function estimation. Eventually, it performs better in sample efficiency and zero-shot generalization than baseline agents, including data augmentation. We observe that the baseline data augmentation methods (RAD and DRAC) sometimes worsen the base model performance. In contrast, our BAE method improves the performance in most tested environments and performs consistently with the base algorithm in other cases. These results show that our method BAE is more robust in performance compared to baseline data augmentation methods. 

The source code of our method is available at \url{https://github.com/masud99r/bae}.

%% file: background.tex
\section{Preliminaries and Problem Settings} \label{sec:background}
\noindent\textbf{Reinforcement Learning}
We assume the task/environment is a Markov Decision Process (MDP) which is denoted by $\mathcal{M} =(\mathcal{S}, \mathcal{A}, \mathcal{P}, \mathcal{R}, \gamma)$, where $\mathcal{S}$ is state space, $\mathcal{A}$ is that action space, $\mathcal{P}$ is the transition probability between states resulted by an action. In this setup, at every timestep $t$, the agent take an action $a_t \in \mathcal{A}$ in a state $s_t \in \mathcal{S}$ and the environment transition to next state $s_{t+1} \in \mathcal{S}$ determined by the transition probability $P(s_{t+1}\vert s_t, a_t)$. 
In a reinforcement learning framework, the goal of the agent is to learn a policy $\pi \in \Pi$ by maximizing the expected return in an MDP, where $\Pi$ is a set of all possible policies and $\pi$ is a single policy which is a mapping from state to action. The policy which achieves the highest expected return is optimal $\pi ^*\in \Pi$.

\noindent\textbf{Policy Gradient}
Proximal Policy Optimization (PPO) \cite{schulman2017proximal}, a type of policy gradient method which achieved tremendous success and is popularly used in many setups because of its effective learning and simple implementation. However, the choice of implementation details might impact the performance of such algorithms in significant ways \cite{andrychowicz2021what,Engstrom2020Implementation,shengyi2022the37implementation}. These implementation details consist of training individual components of the algorithms, such as learning value function, policy function, and advantage estimation. The following is the objective of the PPO \cite{schulman2017proximal}.
\begin{equation}\label{eq:policy_rl}
   \mathcal{L}_\pi = - \mathbb{E}_t[\frac{\pi_\theta(a_t \vert s_t)}{\pi_{\theta_{old}}(a_t \vert s_t)} A_t]
\end{equation}
, where $\pi_\theta(a_t \vert s_t)$ is the probability of choosing action $a_t$ give state $s_t$ at timestep $t$ using the current policy parameterized by $\theta$. On the other hand the $\pi_{\theta_{old}}(a_t \vert s_t)$ refer to the probabilities using an old policy parameterized by previous parameter values $\theta_{old}$. The advantage $A_t$ is an estimation which is the advantage of taking action $a_t$ at $s_t$. 
A popular and effective choice of estimating advantage using a value function is as follows (equation \ref{eq:policy}):
\begin{equation}\label{eq:policy}
    A_t = -V(s_t) + r_t + \gamma r_{r+1} + ... + \gamma^{T-t+1} r_{T-1} + \gamma^{T-t} V(s_T).
\end{equation}
Here $V(s)$ is a value function that gives the average future return under the underlying policy. The first term $V(s_t)$ is the value prediction at timestep $t$, and the rest of the terms except the last term in the equation is the discounted Monte Carlo Estimation which can be computed for a given episode from $t$ to $T-1$ ($T>t$). The last term $V(s_T)$ is the value prediction at state $s_T$. Thus overall, this $A_t$ represents how much the current action $a_t$ is doing compared to the current value prediction. The more the advantage of action, the more the policy should weigh that action. This is done by multiplying $\pi_\theta(a_t \vert s_t)$ with $A_t$. The $\pi_{\theta_{old}}(a_t \vert s_t)$ in Equation \ref{eq:policy} introduced due to importance sampling which allows estimating the advantage from the old policy samples. For details discussion, we refer the reader to \cite{schulman2015trust, schulman2017proximal}.

\noindent\textbf{Value Function Estimation}
An effective value function estimation \cite{schulman2017proximal} is to regress value prediction with an advantage-based return estimation. Here the $V_{error} = |V_{prediction} - V_{Return}|$, where $V_{prediction}$ is the predicted value and the $V_{Return}$ is value computed from the rewards $R = \sum_t r_t$ of sampled trajectories and advantage $A$. Thus, $V_{Return} = A + R$. Note that in this way, both the policy (in equation \ref{eq:policy_rl} and value function is dependent on the advantage estimation. Thus, an accurate advantage estimation should give us lower policy and value losses and thus a better performing policy.

\noindent\textbf{Generalization in RL}
Now we turn attention to the scenarios where different episode varies by confounding features in the observation. Note that these confounders (also called context)impact the reward in the environment; however, they might misguide the agent to think otherwise. Due to nature, the agent might overfit the confounding features and fails to generalize to slightly modified test environments \cite{zhang2018study,Song2020Observational}. 
For a details overview of generalization in reinforcement learning, we refer the reader to the servery papers \cite{kirk2021survey}.

\noindent\textbf{Data augmentation} in various forms has been leveraged \cite{laskin2020reinforcement,raileanu2020automatic}. The general idea is to transform the observation so that the observation's semantic meaning remains the same but the contextual information changes. However, the context information is readily not available, and thus we need to impose various assumptions that certain transformations on the observation $s'=f(s)$ keep the reward semantic. For image-based observation, various image manipulation can be used, such as cropping, rotation, and color-jitter.

\noindent\textbf{Advantage Estimation}
An essential component in policy training is to estimate the advantage. Generalized Advantage Estimation (GAE) \cite{Schulmanetal_ICLR2016} is a useful way to compute advantage, which combines the value function estimate and the Monte Carlo method. However, the data augmentation-based regularization approaches \cite{laskin2020reinforcement, raileanu2020automatic} do not handle the case of advantage estimation. The advantage is estimated using a single trajectory; thus, the computed advantage has a high variance due to the systematic noise in the advantage estimation. Given two similar states (observation), the advantage estimation should be the same. For example, an observation with the same semantic but a different background (red and blue) should have the same advantage if the background is not essential and thus confounded.

%% file: method.tex
\section{Boostrap Advantage Estimation (BAE)}

We proposed to bootstrap the advantage estimation using observation transformation to mitigate the abovementioned issue. Formally, we generate $m$ additional estimation with $m$ transformation. For each such transformation $i$, we compute an estimate as in equation \ref{eq:aug_advantage}.

\begin{equation}\label{eq:aug_advantage}
    A^{(k, i)}_t = -V(f(s_{t+1},v_i)) + r_t + \gamma r_{r+1} + ... + \gamma^{T-t} V(f(s_{t+k},v_i)),
\end{equation}
where $v_0$ refer to no augmentation.
Furthermore, finally, we take the average of all estimates as in equation \ref{eq:bootstrap} to estimate the final advantage estimation for $k$-step return.
\begin{equation}\label{eq:bootstrap}
    A^{(k, b)}_t = \frac{1}{m+1} (A^{(k, 0)}_t + A^{(k, 1)}_t+A^{(k, 2)}_t+..+A^{(k, m)}_t)
\end{equation}
Finally, we can achieve bootstrap advantage estimation of a trajectory of length $T$ by combining several $k$-step returns using exponential-weighted average as in \ref{eq:bae} following the GAE method \cite{Schulmanetal_ICLR2016}.
\begin{equation}\label{eq:bae}
    A^{BAE(\gamma, \lambda)}_t = (1-\lambda)(A^{(1, b)}_t+ \lambda A^{(2, b)}_t  + ... + \lambda^{T-1} A^{(T, b)}_t)
\end{equation}
This $A^{BAE(\gamma, \lambda)}_t$ is used to compute the advantage at state $s_t$ of timestep $t$ in an episode. Note that, our BAE differs from the GAE \cite{Schulmanetal_ICLR2016} in computing the $k$-step return as in equation \ref{eq:bootstrap}.
Algorithm \ref{algo-bae} shows the details step of using our BAE with the PPO-based policy optimization method. 
\begin{algorithm}
    \caption{BAE for Policy Optimization}
    \label{algo-bae}
    \begin{algorithmic}[1]
    \STATE Get transformation function $f(s, v)$ with augmentation type $v$
    \STATE Get PPO for policy optimization RL agent
    \FOR {each iteration}
        \FOR{each environment step} 
             \STATE $a_t \sim \pi_\theta(a_t|s_t)$ 
            \STATE $s_{t+1} \sim P(s_{t+1}|s_t, a_t)$ 
            \STATE $r_t \sim R(s_t, a_t)$ 
            \STATE $\mathcal{B} \xleftarrow{} \mathcal{B} \cup \{(s_t, a_t, r_t, s_{t+1})\} $ 
             \ENDFOR
         \STATE Transform all $s \in \mathcal{B}$ to  get $\mathcal{B}'$ using $v$ augmentation with function $f(s, v)$.
        \STATE Compute Bootstrap Advantage Estimate (BAE) from data $\mathcal{B}$ and $\mathcal{B'}$ using equation \ref{eq:bae}.
        \STATE Perform PPO updates with BAE to optimize for $L_\pi$ as in equation \ref{eq:policy_rl}
       
        \ENDFOR
       
    \end{algorithmic}
    \end{algorithm}
In this paper, we leverage PPO \cite{schulman2017proximal} as the base RL algorithm, which uses generalized advantage estimation (GAE) as the default estimator. In contrast, our method BAE-PPO uses bootstrap advantage estimation (BAE) instead of GAE. The data augmentation baselines RAD and DRAC use base PPO with GAE advantage estimation.
In the experiments we use $m=1$ in equation \ref{eq:bootstrap}. This means we use one data augmentation approach and combine it with original advantage estimation as in equation \ref{eq:bae}.

Note that we do not apply any observation transformation in other parts of the agent objective, and thus equation \ref{eq:policy} remains theoretically and practically sound. Furthermore, we empirically show how our Bootstrap Advantage Estimation leads to a smaller value, policy loss, and performance boost. Finally, we also compared the baseline RAD \cite{laskin2020reinforcement} and DRAC \cite{raileanu2020automatic} and show that our method performs better in many setups. 

%% file: experiment.tex
\section{Experiments} \label{sec:exp}
\subsection{Setup}
\noindent \textbf{Environments}
We experimented with image-based observations with discrete action space and vector-based observations with continuous action space. 

\begin{figure}[!ht]
\centering
\includegraphics[width=0.85\linewidth]{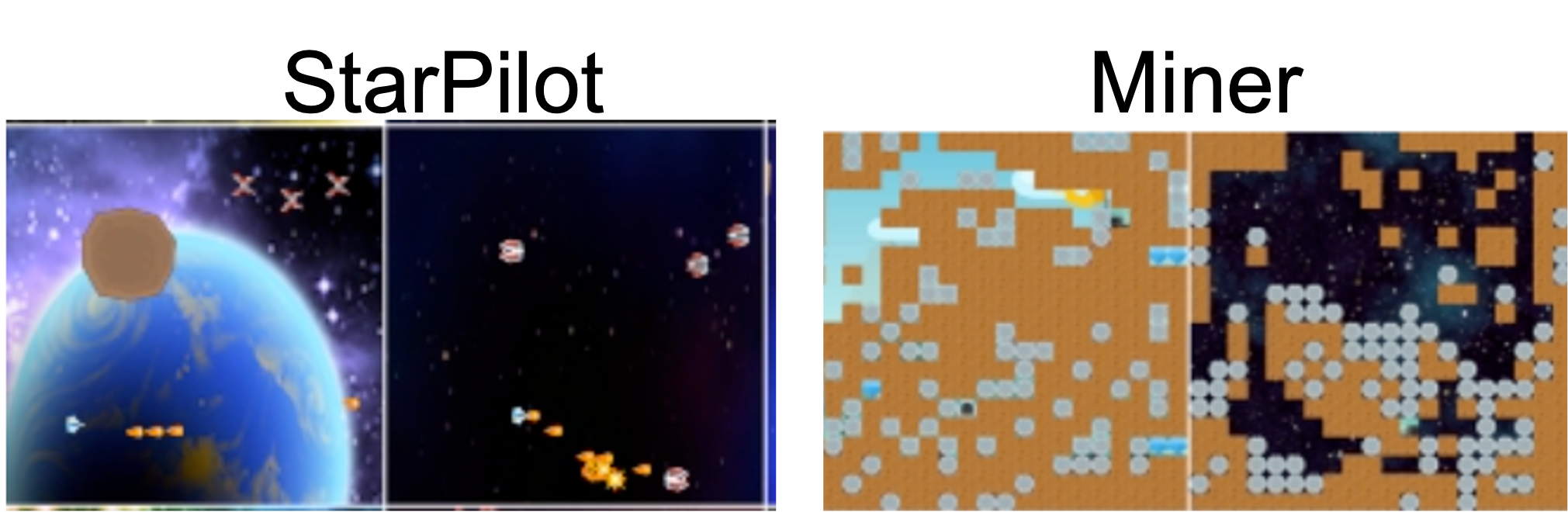} 
\caption{\textbf{Procgen}: Some snapshots of Starpilot and Miner. The environments are generated procedurally, which results in different observations (e.g., background) in each episode. 
}
\centering
\label{fig:procgen_env}
\end{figure}

\noindent \textbf{Procgen} We use Procgen \cite{cobbe2020leveraging}: Starpilot and Miner (Figure \ref{fig:procgen_env}) which use image-based observation and procedural generation to produce challenging game logic that changes episode by episode. This benchmark allows for evaluating both sample efficiency and generalization capacity of RL agents. Each environment has around 100K levels. A subset of levels can be used to train the agent, and then the full distribution, that is, 100K levels, can be used to test the agent's generalization capacity. For our experiment, we use the standard evaluation protocol from \cite{cobbe2020leveraging}; 200 levels of each environment are used for training in the difficulty level \textit{easy}. All the environments have discrete action space of dimension 16. 
Intuitively, during training, the agent has access to a limited number of environment variability (e.g., 200 levels). The trained agent is tested on all the available variabilities, which consist of unseen scenarios. Thus, to master the game, the agent must focus on essential aspects of the state and ignore irrelevant information such as background color.

\begin{figure}[!ht]
\centering
\includegraphics[width=0.99\linewidth]{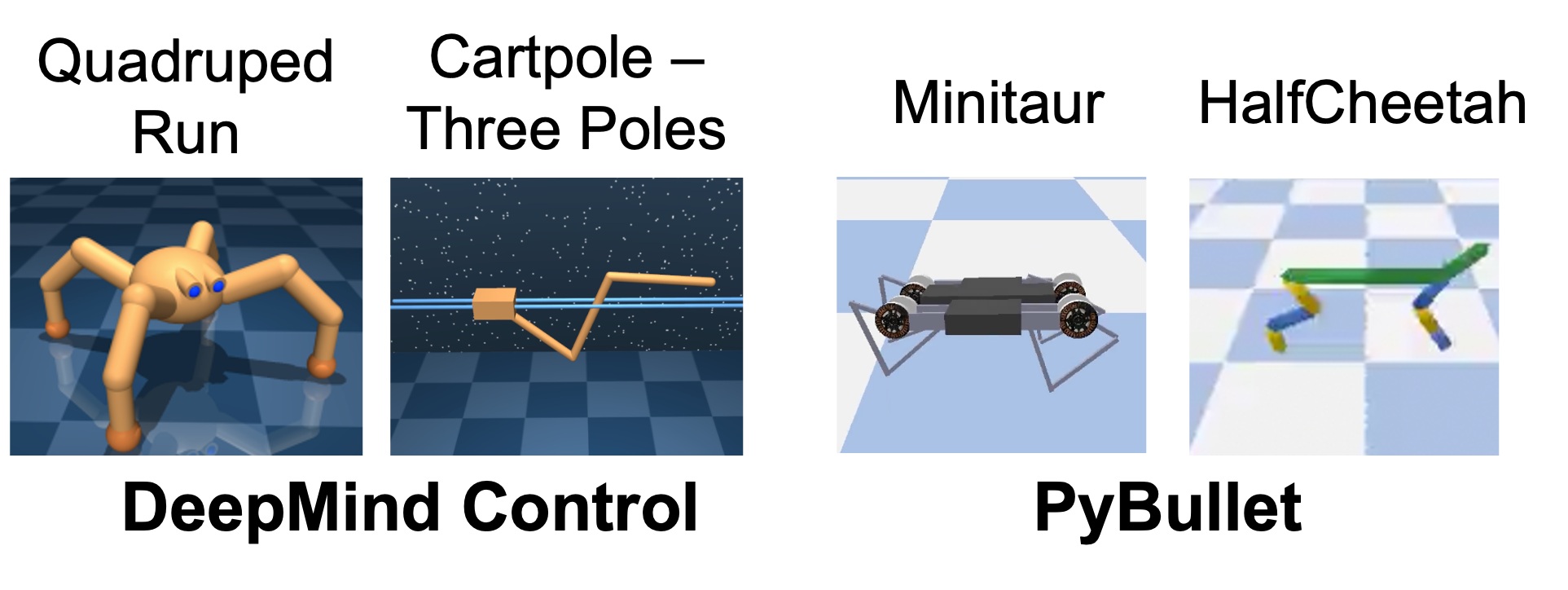} 
\caption{\textbf{[Left]} \textit{Deepmind Control}: Some snapshots of Deepmind Control tasks. \textbf{[Right]} \textit{Pybullet}: Some snapshots of Pybullet Minitaur quadruped and HalfCheetah  environments. These environments contain vector-based state space, and the action space is continuous.}
\centering
\label{fig:dmc_pybullet_env}
\end{figure}

\noindent\textbf{Deepmind Control} We use two environments from \textit{dm\_control} \cite{tunyasuvunakool2020}: Quadruped run, and Cartpole with three poles (Figure \ref{fig:dmc_pybullet_env}). The Quadruped Run has high-dimensional vector observation, and the task is to run as far as possible. On the other hand, the Cartpole variation consists of three procedurally generated poles. The complexity of these environments is suitable for evaluating the data augmentation-based approaches.

\noindent \textbf{Pybullet} We use Pybullet \cite{coumans2021}: Minitaur quadruped and HalfCheetah  environments with vector-based observation. Each observation consists of raw sensory inputs. The Minitaur quadruped is a 4-legged robot, and the task is to travel as long as possible on flat ground. Furthermore, the HalfCheetah is a two-legged robot that can control its movement in 2D, and the task is to travel as much distance as possible. The action spaces are continuous in these environments. Snippets of these environments are in Figure \ref{fig:dmc_pybullet_env}.

\noindent \textbf{Baselines} All agents usage on-policy PPO \cite{schulman2017proximal} as the base policy.
We compare our method with Generalized Advantage Estimation (GAE) \cite{Schulmanetal_ICLR2016} which is referred to as \textbf{GAE-PPO} in our experiments. GAE is shown to perform better compared to other advantage estimation techniques \cite{andrychowicz2021what}.
Moreover, we compare with the data augmentation-based approach uses data augmentation to transform the observation and then uses the transformed observation to train the base policy.
In particular, we compare our method with existing baselines RAD \cite{laskin2020reinforcement} and DRAC \cite{raileanu2020automatic}. RAD, referred to as \textbf{RAD-PPO}, proposes various data augmentation techniques to improve learning from pixel-based observation. DRAC, referred to as \textbf{DRAC-PPO} leverages the data augmentation to regularize the policy and value learning, showing improved performance in policy learning. 
In our method, we replaced the GAE estimation with our proposed Bootstrap Advantage Estimation (BAE), which is referred to as \textbf{BAE-PPO}.

We build all the agents, including our BAE-PPO and baseline, using the implementation available in \cite{shengyi2022the37implementation}. In a PPO-based scenario, many factors have been identified as key in implementing algorithms that impact the performance \cite{andrychowicz2021what, Engstrom2020Implementation}. Thus, we use the same implementation logic for all the baselines and our method for a fair comparison.

\noindent\textbf{Data augmentation} We evaluate \textit{Cutout Color} data augmentation for image-based observation, which performs best in our setup compared to another popularly used Random Crop. Thus, we report Cutout Color data augmentation results for RAD, DRAC, and our BAE. We use the implementation available in RAD \cite{laskin2020reinforcement} for data augmentation.
For the vector-based observation robotic task, we use a random amplitude scale proposed in RAD \cite{laskin2020reinforcement}. This method multiply the observation with a number generated randomly between a range $\alpha$ to $\beta$. We used best performing range $\alpha=0.8$ to $\beta=1.4$ for our BAE method, and a range $\alpha=0.6$ to $\beta=1.2$ for RAD, and DRAC (suggested in RAD\cite{laskin2020reinforcement}).

\noindent\textbf{Implementation and Hyper-parameters}
For the Procgen Starpilot and Miner, we report mean and standard deviation across 3 seeds run following the setup of Procgen paper \cite{cobbe2020leveraging}. We used an Nvidia A100 GPU to run agents with the IMPALA CNN model \cite{espeholt2018impala} on the image observation-based Procgen environments. We use neural networks to represent policy and value functions for vector-based observations. For Deepmind control environments, we report results with 10 random seed runs, and for Pybullet environments, we report results over 5 seeds.
For all experiments, we keep the common hyperparameters the same for a fair comparison. The implementation and hyperparameters are based on \cite{shengyi2022the37implementation,huang2021cleanrl}.
For all results, we report the mean (showed in solid line) and standard deviation (showed in shaded areas) across runs.

\subsection{ Results}
\label{sec:result}
The PPO-based agent's objective consists of value loss and policy loss. The objective is to reduce them and potentially improve the expected return. 
We show that our method BAE reduces the losses and thus learns a better value function and policy than the baselines. We then show how our method performs in the expected return.

\begin{figure}[!tbp]
  \centering
  \begin{minipage}[b]{0.49\columnwidth}
    \includegraphics[width=0.99\textwidth]{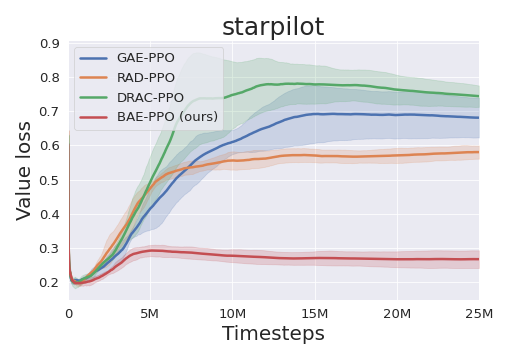}
  \end{minipage}
  \hfill
  \begin{minipage}[b]{0.49\columnwidth}
    \includegraphics[width=0.99\textwidth]{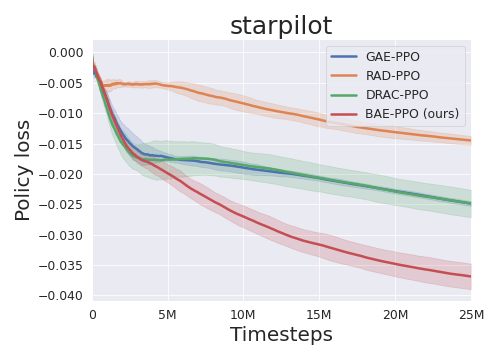}
  \end{minipage}
  \hfill
  \caption{\textit{Starpilot Env.} Training time \textbf{policy} and \textbf{value} loss [\textbf{lower is better}]. Our method achieves lowest value and policy losses than the base algorithm (GAE-PPO) and data augmentation baselines (RAD and DRAC).
  }
  \label{fig:starpilot_loss}
\end{figure}

\noindent\textbf{Procgen Results} Figure \ref{fig:starpilot_loss} shows value and policy loss during policy training on Procgen \textbf{Starpilot} environments. As the training progresses, the loss of our method BAE reduces drastically compared to the baselines. These results show the sign of the effectiveness of our method in reducing the agent's losses. Note that the advantage estimation is used to train both the value function and the policy; thus, better estimation of advantages should generally give better value and policy. In this sense, our method shows empirical evidence that it can help better advantage estimation.

\begin{figure}[!tbp]
  \centering
  \begin{minipage}[b]{0.49\columnwidth}
    \includegraphics[width=0.99\textwidth]{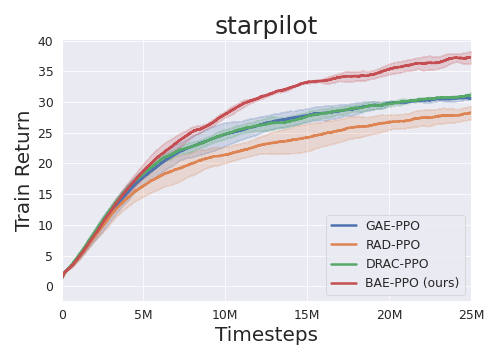}
  \end{minipage}
  \hfill
  \begin{minipage}[b]{0.49\columnwidth}
    \includegraphics[width=0.99\textwidth]{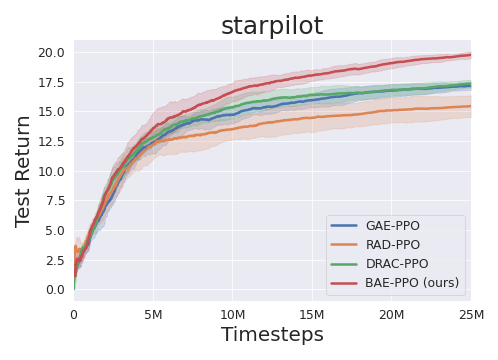}
  \end{minipage}
  \hfill
  \caption{\textit{Starpilot Env.} \textbf{Sample efficiency} performance measured in train time return. We see our method BAE-PPO achieves higher returns where DRAC does not improve the base agent's (GAE-PPO) performance. RAD slightly worsens the performance of the base agent. [\textbf{Right}] \textbf{Generalization} performance measured in test time return. We see a similar trend and observe that our method performs the best.
  }
  \label{fig:starpilot_return}
\end{figure}

We observe that in the Starpilot environment, the success in policy and value loss translates to the final return. In Figure \ref{fig:starpilot_return} we see that our method (BAE-PPO) shows improved sample efficiency (Train Return) and generalization capacity compared to the baseline GAE-PPO, RAD-PPO, and DRAC-PPO. 

Note that the RAD-PPO performance worsens the performance of the base GAE-PPO algorithms. This result is consistent with the findings in \cite{raileanu2020automatic}, and it shows that naive data augmentation can be detrimental to performance. Furthermore, we observe a similar trend for the value and policy loss results.

\begin{figure}[!tbp]
  \centering
  \begin{minipage}[b]{0.49\columnwidth}
    \includegraphics[width=0.99\textwidth]{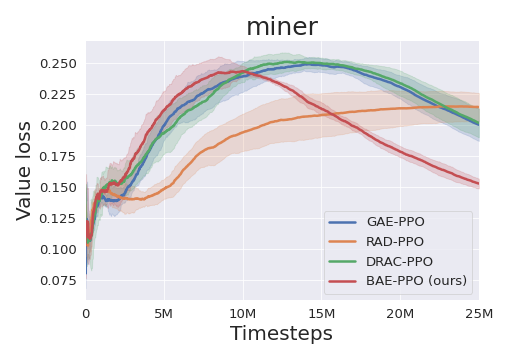}
  \end{minipage}
  \hfill
  \begin{minipage}[b]{0.49\columnwidth}
    \includegraphics[width=0.99\textwidth]{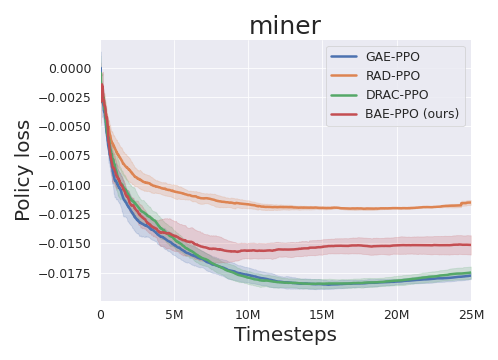}
  \end{minipage}
  \hfill 
  \caption{\textit{Miner Env.} Training time \textbf{policy} and \textbf{value} loss [\textbf{lower is better}]. The value loss of BAE eventually become lower. Losses of RAD increases compared to base PPO.
  }
  \label{fig:miner_loss}
\end{figure}

We observe a similar performance trend in the Procgen \textbf{Miner} environment. In Figure \ref{fig:miner_loss}, we see that in our method, BAE shows a smaller value loss eventually despite being higher at the beginning compared to the baselines. GAE and DRAC show slightly lower values in the policy loss plot than BAE. However, both show smaller policy losses in general. 

\begin{figure}[!tbp]
  \centering
  \begin{minipage}[b]{0.49\columnwidth}
    \includegraphics[width=0.99\textwidth]{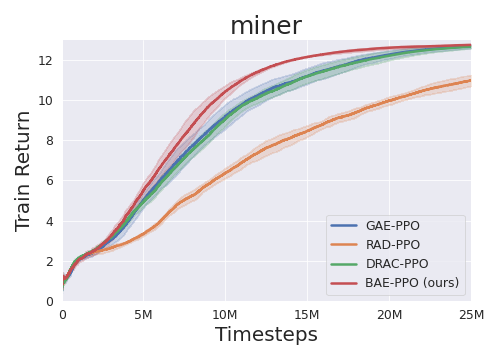}
  \end{minipage}
  \hfill
  \begin{minipage}[b]{0.49\columnwidth}
    \includegraphics[width=0.99\textwidth]{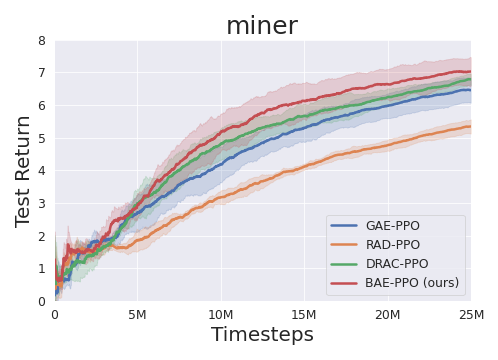}
  \end{minipage}
  \hfill
  \caption{\textit{Miner Env.} [\textbf{Left}] \textbf{Sample efficiency} performance measured in train time return and [\textbf{Right}] \textbf{Generalization} performance measured in test time return. Overall, we see our method BAE-PPO shows consistence improvement over baselines.
  }
  \label{fig:miner_return}
\end{figure}

In performance measure (Figure \ref{fig:miner_return}), we see that our method shows better performance throughout the training than the baseline in both sample efficiency and generalization. The performance difference is consistent across timestep. Similar to Starpilot, RAD also performs worse compared to base GAE-PPO.

We further evaluate our method on vector-based state space and continuous robotic tasks: Deepmind control and Pybullet. Note that the setup of these benchmarks are different from Procgen's train-test setup, and here we evaluate how our agents can perform in high-dimensional states and procedurally generated task. Thus we can only report returns during training.

\begin{figure}[!tbp]
  \centering
  \begin{minipage}[b]{0.49\columnwidth}
    \includegraphics[width=0.99\textwidth]{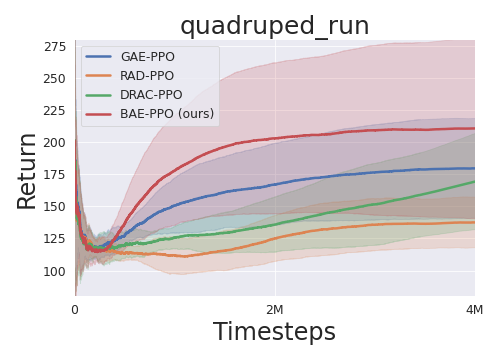}
  \end{minipage}
  \hfill
  \begin{minipage}[b]{0.49\columnwidth}
    \includegraphics[width=0.99\textwidth]{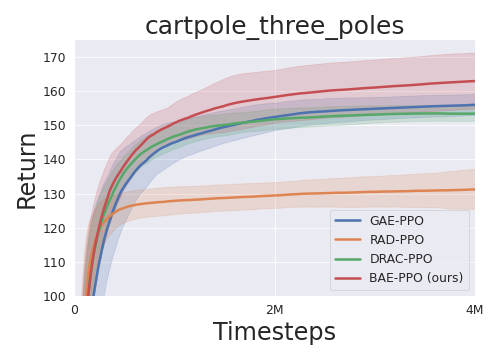}
  \end{minipage}
  \hfill 
  \caption{[\textbf{Left}] Performance on Quadruped run environments. Our method BAE-PPO shows higher mean returns compared to all other agents. Data augmentation baseline RAD and DRAC worsen the performance of the base agent (GAE-PPO). [\textbf{Right}] Our method BAE consistently achieves a higher mean where DRAC fails to improve upon the base agent, and RAD worsens the base performance.
  }
  \label{fig:dmc_return}
\end{figure}

\noindent\textbf{Deepmind Control Results} Figure \ref{fig:dmc_return} shows performance compariosn on Quadruped run and Cartpole - Three Poles environments.
We observe that our method BAE achieves the best performance in both environments. On the other hand, the baseline RAD severely worsens the base agent's (GAE-PPO) performance in both environments. Another baseline, DRAC, worsens the base agents' performance in Quadruped Run and fails to improve performance in the Cartpole Three Poles environment. These results show that properly using the augmentation in policy learning can lead to strong performance. In this case, all data augmentation agents (RAD, DRAC, and BAE) use the same \textit{random amplitude modulation} augmentation. However, using this augmentation in advantage computation, our method BAE shows a substantial performance boost. On the other hand, other baselines, RAD and DRAC, worsen the performance (Quadruped Run).

\begin{figure}[!tbp]
  \centering
  \begin{minipage}[b]{0.49\columnwidth}
    \includegraphics[width=0.99\textwidth]{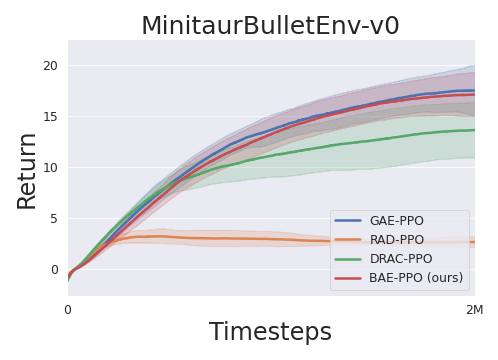}
  \end{minipage}
  \hfill
  \begin{minipage}[b]{0.49\columnwidth}
    \includegraphics[width=0.99\textwidth]{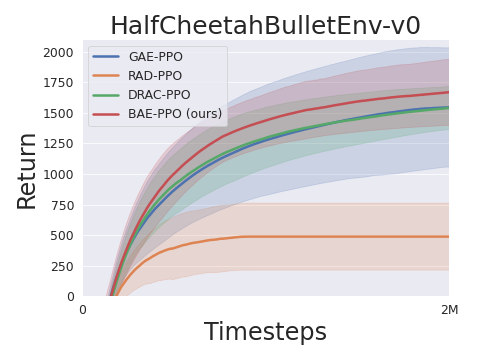}
  \end{minipage}
  \hfill
  \caption{Performance on Minitaur [\textbf{Left}] and HalfCheetah [\textbf{Right}] environments. Our method shows better or similar performance compared to the base agent (GAE-PPO), where the data augmentation baselines sometimes worsen the base performance.
  }
  \label{fig:pybullet_return}
\end{figure}

\noindent\textbf{Pybullet Results}
In Figure \ref{fig:pybullet_return}, for the HalfCheetah environment, we see that our method BAE performs better than base agent GAE-PPO and other data augmentation baselines RAD and DRAC. On the other hand, in Minitaur, the data augmentation baseline DRAC and RAD worsen the base agent's (GAE-PPO) performance where BAE can maintain the base performance.

Overall, these results show the robustness of our method in performance compared to baseline data augmentation methods. Therefore, the proposed augmented observation is expected not to worsen the base performance. However, in our experiments, we observe that the RAD and DRAC barely match the base agent's (GAE-PPO) results and sometimes worsen the performance. These variabilities in performance hinder the widespread adaptation of these methods. In contrast,  our method BAE shows a consistent performance across various tasks without reducing the base agent's performance. 

%% file: relatedwork.tex
\section{Related Work} 
\label{sec:related_work}
\noindent\textbf{Advantage Estimation}. The baseline has been leveraged to reduce variance in policy gradient update \cite{williams1992simple,peters2008reinforcement,wu2018variance}. Furthermore, effective modification of such methods is the use of advantage estimation. This method is commonly used in policy optimization and enjoys strong empirical success \cite{andrychowicz2021what}, especially the Generalized Advantage Estimation (GAE) \cite{Schulmanetal_ICLR2016} method. In contrast to GAE, our method BAE leverages data augmentation and incorporates advantage computation across various semantically similar states. Empirically we observe that our method of computing advantage can be beneficial over GAE, especially in high-dimensional and procedural generated environments.

\noindent\textbf{Data augmentation}. Data augmentation has been demonstrated to be an effective and efficient approaches to improve performance \cite{cobbe2019quantifying,laskin2020reinforcement,raileanu2020automatic}. Other methods proposed to improve generalization which includes regularization \cite{igl2019generalization}, and style-transfer \cite{rahman2022bootstrap}.
Depending on how the augmented observation is used, the method can be different; for example, RAD \cite{laskin2020reinforcement} and DRAC \cite{raileanu2020automatic}. In contrast to these methods, our method incorporates data augmentation into advantage estimation, which shows better empirical performance compared to these methods (RAD and DRAC). 

%% file: conclusion.tex
\section{Conclusion}
In this paper, we propose a data augmentation-based advantage estimation method for policy optimization. Our Bootstrap advantage estimation (BAE) method replaces the GAE method in policy gradient-based algorithms. We demonstrated the effectiveness of our method on PPO algorithms. Furthermore, we evaluated our methods on both image-based observation space with discrete action space and vector-based observation with continuous action space (Procgen, Deepmind Control, and Pybullet). Our BAE method showed better performance in various environment setups than GAE. Furthermore, our method performs better than two existing data augmentation techniques (RAD and DRAC).